\documentclass[letterpaper, 10 pt, conference]{ieeeconf}  

\IEEEoverridecommandlockouts                              

\usepackage{amsmath}
\usepackage{amssymb} 
\usepackage{graphicx}
\usepackage{tabularx}
\usepackage{booktabs}
\usepackage{dcolumn}
\usepackage{multirow}
\usepackage{url}
\usepackage[skip=2pt,font=small]{caption}
\usepackage{authblk}
\usepackage{dblfloatfix}
\makeatletter
\let\NAT@parse\undefined
\makeatother
\usepackage[numbers,sort&compress,sectionbib]{natbib}
\usepackage{diagbox}
\usepackage{color}
\usepackage[colorlinks=true,urlcolor=blue,citecolor=black]{hyperref}

\newcolumntype{C}{>{\centering\arraybackslash}X}
\long\def\ignore#1{}
\long\def\ignorethis#1{}
\def\idigit{\phantom{0}} 

\def\placeholder#1#2{\fbox{\vbox to #2{\vss\hbox to #1{\hss~}\vss}}}

\title{\LARGE \bf
High-Precision Localization Using Ground Texture
}

\makeatother
\author{Linguang Zhang \quad Adam Finkelstein \quad Szymon Rusinkiewicz 
  }
\affil{Princeton University}
\begin{document}

\maketitle
\thispagestyle{empty}
\pagestyle{empty}


\begin{abstract}
Location-aware applications play an increasingly critical role in everyday
life.  However, satellite-based localization (e.g., GPS)
has limited accuracy and can be unusable in
dense urban areas and indoors.  We introduce an image-based global
localization system that is accurate to a few millimeters and performs
reliable localization both indoors and outside.  The key idea is to capture
and index distinctive local keypoints in ground textures.  This is based on
the observation that ground textures including wood, carpet, tile,
concrete, and asphalt may look random and homogeneous, but all contain
cracks, scratches, or unique arrangements of fibers.  These
imperfections are persistent, and can serve as local features.  Our system
incorporates a downward-facing camera to capture the fine texture of the
ground, together with an image processing pipeline that locates the
captured texture patch in a compact database constructed offline.  We
demonstrate the capability of our system to robustly, accurately, and
quickly locate test images on various types of outdoor and indoor ground
surfaces.
This paper contains a supplementary video. 
All datasets and code are available online at 
\href{http://microgps.cs.princeton.edu}{microgps.cs.princeton.edu}.
\end{abstract}
\section{Introduction}
The Global Positioning System (GPS) receiver has become an essential 
component of both hand-held mobile devices and vehicles of all types.
Applications of GPS, however, are constrained by a number of
known limitations.  A GPS receiver must have access to unobstructed lines
of sight to a minimum of four satellites, and obscured satellites
significantly jeopardize localization quality.  Indoors, a GPS receiver
either is slow to obtain a fix, or more likely does not work at all.  
Even outdoors, under optimal circumstances, accuracy is limited to a
few meters (or perhaps a meter with modern SBAS systems).  These
limitations make GPS insufficient for fine-positioning applications such as
guiding a car to a precise location in a parking lot, or guiding a
robot within an indoor room or warehouse. 
To overcome the robustness and accuracy limitations of GPS, alternative
localization technologies have been proposed, which are either less
accurate than GPS (e.g., triangulation of cellphone towers and WiFi
hotspots), or expensive and cumbersome to deploy (e.g., RFID
localization or special-purpose sensors embedded in the environment). 
Inertial navigation and odometry, which are often used in robotics for
fine-positioning tasks, require a known initial position, drift over time,
and lose track (requiring manual re-initialization) when the device is
powered off.


\begin{figure}
  \centerline{\includegraphics[width=\hsize]{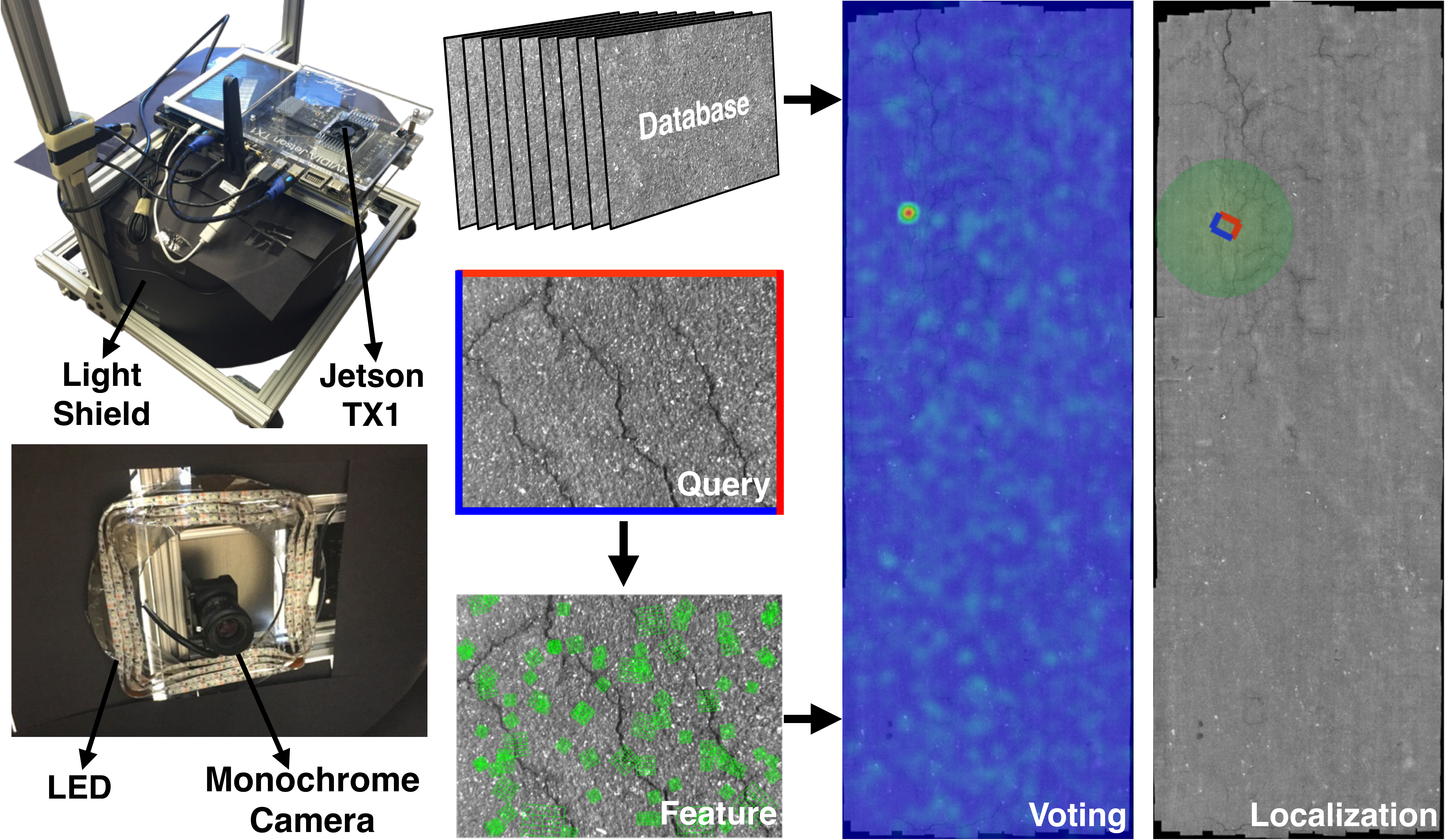}}
  \caption{System overview.
  Our test robot features an NVIDIA Jetson TX1 development board, which
  controls a Point Grey mono camera. A light shield and ring of LEDs around
  the camera provide controlled lighting. We first capture database
  images as a preprocess, and stitch them into a globally consistent
  map. Later, to locate the robot, we extract features from a query
  image, and they vote for potential image poses in the map. A peak in
  the voting map determines inlier features, from which we recover the
  pose of the query image.
  }
  \label{fig:teaser}
\end{figure}

This paper proposes a system 
that provides millimeter-scale localization,
both indoors and outside on land.
The key observation behind our approach is that seemingly-random ground
textures exhibit distinctive features that, in combination, provide a means
for unique identification.  Even apparently homogeneous 
surfaces contain small imperfections~-- cracks, scratches, or even a
particular arrangement of carpet fibers~--
that are persistently identifiable as local features.  
While a single feature is not likely
to be unique over a large area, the spatial relationship among a group of
such features in a small region is likely to be distinctive, at least
up to the uncertainty achievable with coarse localization methods such as GPS
or WiFi triangulation.
Inspired by this observation, we construct a system
called Micro-GPS that includes a
downward-facing camera to capture fine-scale ground textures, and an image
processing unit capable of locating that texture patch
in a pre-constructed compact database within a few hundred milliseconds.

The use of image features for precise localization has a rich history,
including works such as Photo Tourism~\cite{Snavely2006} and
Computational Re-Photography~\cite{Bae2010}.  Thus, a main contribution of
our work is determining how some of the algorithms used for feature
detection and matching in ``natural'' images, as used by previous work,
can be adapted for ``texture-like'' images of the ground. 
In searching for a robust combination of such methods,
we exploit two key advantages of ground-texture images.  
First, the ground can be photographed from much
closer range than typical features in the environment, leading to an
order-of-magnitude improvement in precision.  Second, the statistics of
texture-like images lead to a greater density of features, leading to
greater robustness over time.

Our system consists of two phases: 
an offline database construction phase, 
and an online localization phase (Figure \ref{fig:teaser}).
We begin by collecting
ground texture images and aligning them using global pose optimization.
We extract local features (keypoints) and store them in a database, which
is subsequently compressed to a manageable size.  
For localization, we find keypoints in a query image
and search the database for candidate matches using
approximate nearest neighbor matching.
Because it is common for more than 90\% of the matches to be spurious,
we use voting to reject outliers, 
based on the observation that inlier matches will vote
for a consistent location whereas outliers distribute their votes randomly.
Finally, we use the remaining inlier matches to precisely calculate
the location of the query image.

The major contributions of this paper are:
\begin{itemize}

\item Describing a low-cost global localization system based on ground
textures and making relevant code and instructions available for reproduction.

\item Capturing and making available datasets of seven indoor and outdoor
ground textures.



\item Investigating the design decisions necessary for practical matching
in texture-like images, as opposed to natural images.  This includes the
choice of descriptor, strategies for reducing storage costs, and a robust
voting procedure that can find inliers with high reliability.

\item Demonstrating a real-world application of precise localization: a
robot that uses Micro-GPS to record a path and then follow
it with sub-centimeter accuracy.





\end{itemize}

The ability to localize a vehicle or robot precisely has the
potential for far-reaching applications.  A car could accurately park
(or guide the driver to do so) in any location it recognizes from
before, avoiding obstacles mere centimeters away.  A continuously-updated
map of potholes could be used to guide drivers to turn slightly
to avoid them.  The technology applies equally well to vehicles
smaller than cars, such as Segways, electric wheelchairs, and mobility
scooters for the elderly or disabled, any of which could be guided to
precise locations or around hard-to-see obstacles.  Indoor
applications include guidance of warehouse robots and precise control over
assistive robotics in the home.

\ignorethis{ Mostly covered by prev paragraph:
We believe that a new generation of groundbreaking location-aware
applications can be enabled by a Global Micro-Positioning System that is
hundreds of times more precise than GPS and is reliable both indoors and
out.  Whether deployed on board a car, truck, wheelchair, or autonomous
robot, we aim to achieve millimeter-scale localization for most places on
land:  the roads, paths, garages, warehouses, rooms, and hallways of the
world.  The hardware required to build this system is inexpensive and
already available.  The coverage of the system is not limited by satellite
visibility, and can be continuously increased and updated.
}

\begin{figure*}[t]
  \small
  \centering
  \includegraphics[width=\hsize]{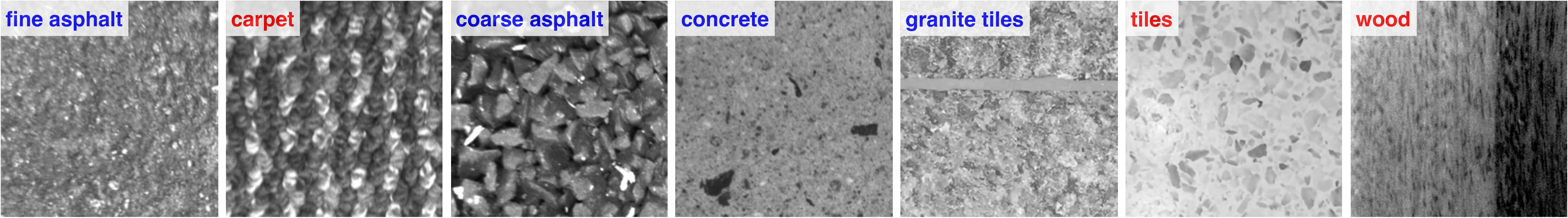}
  \caption{Example texture patches cropped from our dataset.
  \textcolor{blue}{Outdoor} and \textcolor{red}{indoor} 
  textures are marked in blue and red respectively.}
  \label{fig:example_textures}
\end{figure*}
\section{Related Work}

{\noindent \bf Textures for Tracking and Localization:}
Textures such carpet, wood grain, concrete or asphalt all have bumps,
grooves, and variations in color from location to location, and we
typically use the overall pattern or \emph{statistics} of this variation to
recognize a particular material. Indeed computer-based modeling and
recognition of textures traditionally proceeded along statistical
lines~\cite{Dana:1999,Leung:2001}. Moreover, researchers have successfully
synthesized new texture by example using
parametric~\cite{heeger1995pyramid} and
non-parametric~\cite{efros1999texture} models.
However, when we study the \emph{particular} arrangement of
bumps and variations present at any location in real-world textures,
we find that it is unlikely to be repeated elsewhere.

\citet{kelly2007field} introduce a warehouse automation
system in which a downward facing camera installed on each robot is used to
help track the robot.  They observe that ground surfaces usually
exhibit cracks and scratches, and it is possible to track
the motion of the camera over a pre-constructed visual map.  This work,
however, still assumes a known initial location and surface textures
are leveraged only for pairwise (\emph{local}) frame matching, 
much as is done in an optical mouse.
Other similar systems~\cite{fang2009ground, kozak2016ranger} align the
test frame with a small set of map frames determined either by an odometry
or the most recent successful estimation. 
In contrast, our approach performs \emph{global} localization, which could
be used to initialize tracking systems such as these.

\citet{clarkson2009fingerprinting} demonstrate that
seemingly-random textures can provide a means for unique identification.
The authors observe that the fine-scale variations in the fibers of a piece
of paper can be used to compute a ``fingerprint" that uniquely identifies
the piece of paper.  Our work demonstrates that ground textures, including
man-made ones such as carpet, share similar properties at sufficiently fine
scales, and thus may be used for localization.

{\noindent \bf Relocalization:}
Structure from motion allows reconstruction of a large scale 3D point cloud
offline, but relocating a newly captured image in the reconstructed point
cloud without any initial guess about the camera position is challenging.
Previous works explore direct 2D-to-3D matching \cite{sattler2011fast} to
estimate the 6~DoF pose of a photo with respect to a reconstructed point
cloud.  \citet{li2012worldwide} propose a method to leverage a
co-occurrence prior for RANSAC and achieve relocalization on a larger
georegistered 3D point cloud within a few seconds.
Relocalization is an essential module of modern SLAM systems, such as
ORB-SLAM~\cite{mur2015orb}, which uses a bag-of-words model for matching.
\citet{kendall2015posenet} train a convolutional neural network (PoseNet)
to regress the input RGB image to the 6-DoF camera pose. 
Researchers have also explored using skylines from omni-images to perform
relocalization~\cite{ramalingam2009geolocalization}.

All the above approaches, except PoseNet, involve large-scale feature
matching, which quickly becomes a bottleneck because of the number of
false matches.  To speed up feature matching, more
compact models can be constructed by selecting a subset of stable 3D points
from the original models \cite{li2010location,cao2014minimal}.  An
effective approach to handle a high outlier ratio is
voting~\cite{zeisl2015camera}.  This has also proven successful in the
field of image retrieval, where spatial verification is commonly applied to
rerank the retrieved list of images, and variants of Hough voting have been
proposed to improve efficiency and
robustness~\cite{avrithis2014hough,wu2015adaptive,schonbergervote}.  With
more sensors available, one can utilize the gravity direction
\cite{svarm2016city} as an additional constraint. \citet{baatz2010handling}
leverage both gravity direction and a 3D scene model to rectify images,
transforming the 6-DOF pose estimation problem into a 3-DOF problem. 

Mobile devices are ideal deployment platforms for a relocalization system. 
\citet{lim2015real} achieve localization on a micro aerial
vehicle at interactive framerates by distributing feature extraction over
multiple frames.  \citet{middelberg2014scalable}
achieve real-time performance by combining online camera tracking and an
external localization server.
\citet{irschara2009structure}
and \citet{wendel2011natural} demonstrate that GPUs,
which are now widely available on mobile processors, can be used to accelerate
localization.


Almost all of the above relocalization techniques rely on landmarks, such
as buildings, that are normally positioned at least a few meters from the
camera.  This distance, combined with finite image resolution and inherent
uncertainty in camera intrinsics, means that even a small error in feature
localization results in a large uncertainty in estimated camera pose.  This
inaccuracy can be ameliorated by increasing the field of view of the
camera~\cite{schonbein2014omnidirectional,arth2011real}, because as more
features are detected, more constraints can be introduced to improve pose
estimation.  Further uncertainty comes from the ambiguity in landmark
identification, since it is not unusual to find buildings or parts of
buildings (such as windows) that appear the same from a significant
distance.  Moreover, natural images used by the above systems suffer from
perspective foreshortening, which brings difficulties to feature matching. 
Many features are not ``time-invariant'' in highly dynamic scenes. Thus, it
is necessary to update the database frequently.  Finally, these systems can
be affected by changes in lighting.  In contrast with these systems, our
work positions the camera close to the texture being imaged and uses
controlled lighting, leading to higher precision and robustness.

\section{System}

\subsection{Mapping}
{\noindent \bf Hardware Setup and Data Collection:}
Our imaging system consists of a Point Grey CM3 grayscale camera pointed
downwards at the ground (Figure \ref{fig:teaser}, left).  A shield blocks
ambient light around the portion of the ground imaged by the camera, and a
set of LED lights arranged symmetrically around the lens provides
rotation-invariant illumination.  The distance from the camera to the
ground is set to 260~mm for most types of textures we have experimented
with.  Our system is insensitive to this distance, as long as a
sufficient number of features can be detected.
The camera output is processed by an NVIDIA Jetson TX1 development board.
Our prototype has the camera and development board mounted on a mobile
cart, which may be moved manually or can be driven with a pair of
computer-controlled motorized wheels.  The latter capability is used for
the ``automatic path following'' demonstration described in
Section~\ref{sec:follow}.  For initial data capture, however, we manually
move the cart in a zig-zag path to ensure that an area can be fully
covered. This process, while is easily mastered by non-experts, could be
automated by putting more engineering effort or even through crowd-sourcing
when there are more users.

{\noindent \bf Image Stitching:}
To construct a global map, we assume the that surface is locally planar,
which is true even for most outdoor surfaces. Our image stitching pipeline
consists of frame-to-frame matching followed by global optimization,
leveraging extensive loop closures provided by the zig-zag path. Since the
computation becomes significantly more expensive as the area grows, we
split a large area into several regions (which we reconstruct separately) 
and link the already-reconstructed regions. This allows us to quickly map larger
areas with decent quality. Figure~\ref{fig:teaser}, right shows the ``asphalt''
dataset, which covers 19.76$\text{m}^2$ in high detail.


{\noindent \bf Datasets:}
We have experimented with a variety of both indoor and outdoor datasets,
covering ground types ranging from ordered (carpet) to highly stochastic
(granite), and including both the presence (concrete) and absence (asphalt)
of visible large-scale variation. We have also captured test images 
for the datasets on a different day (to allow perturbations to the
ground surfaces) to evaluate our system. 
Figure~\ref{fig:example_textures} shows example patches from our dataset. 
We will make these datasets,
together with databases of SIFT features and test-image sequences,
available to the research community.  

{\noindent \bf Database Construction:}
The final stage in building a map is extracting a set of features from the
images and constructing a data structure for efficiently locating them. 
This step involves some key decisions, which we evaluate in 
Section~\ref{sec:evaluation}.  Here we only describe our actual implementation.
We use the SIFT scale-space DoG detector and gradient orientation histogram
descriptor~\cite{lowe2004distinctive}, since we have found it to have high
robustness and (with its GPU implementation \cite{wu2007siftgpu}) reasonable
computational time. For each image in the map, we typically find 1000 to
2000 SIFT keypoints, and randomly select 50 of them to be stored in the
database. This limits the size of the database itself, as well as the data
structures used for accelerating nearest-neighbor queries. We choose random
selection after observing that features with higher DoG response are not
necessarily highly repeatable features:  they are just as likely to be due
to noise, dust, etc. 
To further speed up computation and reduce the size of the database, we apply
PCA~\cite{pearson1901liii} to the set of SIFT descriptors and project each descriptor onto the top
$k$ principal components.  As described in Section~\ref{sec:evaluation}, for
good accuracy we typically use $k=8$ or $k=16$ in our implementation, and
there is minimal cost to using a ``universal'' PCA basis constructed from
a variety of textures, rather than a per-texture basis.


One of the major advantages of our system is that the height of the camera
is fixed, so that the scale of a particular feature is also fixed.  This
means that when searching for a feature with scale $s$ in the database, we
only need to check features with scale $s$ as well.  In practice, to allow
some inconsistency, we quantize scale into 10 buckets and divide the
database into 10 groups based on scale.  Then we build a search index for
each group using FLANN~\cite{muja2009fast}.  During testing,
given a feature with scale $s$, we only need to search for the nearest
neighbor in the group to which $s$ belongs.

\subsection{Localization}
\begin{figure}
  \centering
  \includegraphics[width=\hsize]{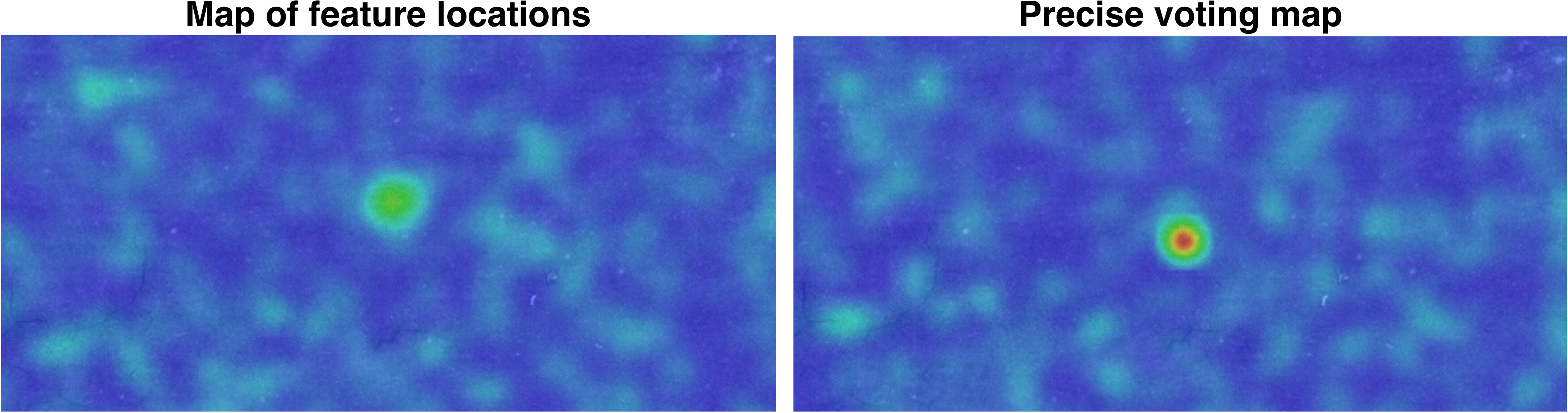} \\
  \caption{Left: A simple strategy would be to vote for the locations of
    matched features.  
    Right: The precise voting leads to a more defined local maximum.}
  \label{fig:precise_voting}
\end{figure}

\label{sec:online_localization}
The input to our online localization phase is a single image.  We assume
that the height of the camera above the ground is the same as during
mapping (or that the ratio of heights is known), so that the image scale is
consistent with the images in the database. 

{\noindent \bf Feature Computation and Matching:}
We first extract SIFT features from the
test image and project onto $k$ principal components,
as in database construction.  
For each descriptor, we search for the nearest neighbor using the
pre-build search index for the appropriate scale.

{\noindent \bf Precise Voting:}
Recall that we only keep 50 features per database image, so
only a small subset of features will have a correct match in the database.
Finding this small set of inliers is challenging, since
methods such as RANSAC work poorly if outliers greatly outnumber
inliers.

We instead adopt a voting approach
based on the observation that, due to the randomness of
ground textures, false matches are usually evenly distributed in the map. 
Fortunately, since true matches usually come from one or two images, they
are concentrated in a small cluster.  Figure \ref{fig:precise_voting}, left,
shows a heat map of feature matches in a database, with red indicating high
density, green intermediate, and blue indicating low density.  While we are
able to build a system based on this principle, the correct features are
distributed throughout the entire area corresponding to the test image. 
This leads to poor robustness, because there is only a moderately-high
density of votes in the map near the location of the test image.
The solution is to concentrate
the votes: we want all of the true features to vote for the \emph{same}
point in the map, leading to a much greater difference between the peak
corresponding to the true location and the background density of outliers.

In particular, each feature casts a vote for \emph{the origin of the test
image} by assuming that nearest neighbors are true matches.  Denote a
feature extracted from the test image as $f_t$ and its nearest neighbor in
the database as $f_d$.  If the feature pair $\{f_t, f_d\}$ is a true match,
we can compute the pose of the test image $T$ in world coordinates, denoted
$[R|t]_T^W$, by composing the pose of $f_d$ in world coordinates and the
pose of $f_t$ in the test image:
\begin{equation}
[R|t]_T^W = [R|t]_{f_d}^W \; [R|t]_{f_t}^{f_d} \; [R|t]_T^{f_t},
\end{equation}
where $[R|t]_{f_t}^{f_d}$ is assumed to be the identity.  We then vote for
the location of the origin of the test image, which is the translational
component of $[R|t]_T^W$. 

Using this strategy, implemented via voting on a relatively fine spatial
grid with each cell set to 50$\times$50 pixels, we find a much tighter
peak of votes relative to the uniform background of outliers, as shown in
Figure \ref{fig:precise_voting}, right. After voting, the cell with the
highest score is very likely to contain the true origin of the test image.
We select all of the features in that peak as likely inliers, and perform
RANSAC just on them to obtain a final estimate of the pose of the image.
\begin{table}[t]
  \centering
  \caption{Performance of Micro-GPS. From left to right: 
  texture type, elapsed time between capture of database and test sequence,
  number of test frames, and
  success rates using 8- and 16-dimensional descriptors.}
  \label{tab:microgps}
  \begin{tabularx}{\hsize}{Xcccc}
    \toprule
    \bf Texture & \bf Elapsed & \bf \# frames & \bf Rate-8 & \bf Rate-16    \\
    \midrule
    fine asphalt   & 16 days       & \idigit651   & 76.04\%   & 95.24\%  \\  
    carpet         & 30 days       & 1179         & 99.49\%   & 99.92\%  \\ 
    coarse asphalt & 17 days       & \idigit771   & 97.54\%   & 99.09\%  \\ 
    concrete       & 26 days       & \idigit797   & 83.31\%   & 93.35\%  \\ 
    granite tiles  & \idigit6 days & \idigit862   & 79.47\%   & 94.43\%  \\ 
    tiles          & 18 days       & 1621         & 93.83\%   & 98.40\%  \\ 
    wood           & \idigit0 days & \idigit311   & 59.48\%   & 77.49\%  \\
    \bottomrule
  \end{tabularx}

\end{table}

\section{Evaluation}
\label{sec:evaluation}

In order to evaluate the accuracy and robustness of a localization system,
a typical approach would be to obtain ground-truth location and pose using
a precise external measurement setup.  However, this is impractical in our
case due to the large areas mapped and the precision with which we are
attempting to localize.  Moreover, we are more interested in
\emph{repeatability}, rather than absolute accuracy, given that most of the
applications we envision will involve going to (or avoiding) previously-mapped
locations.

We therefore adopt an evaluation methodology based on comparing the query
image against an image captured during mapping.  Using the pose predicted
by Micro-GPS, we find the closest image in the database, and compute
feature correspondences (using all SIFT features in the image, not just
the features stored in the database).  If there are insufficiently many
correspondences, we mark the localization result as a failure.  We then
compute a best-fit relative pose using those features.  If the pose differs
by more than 30 pixels (4.8~mm) in translation or 1.5${}^\circ$ in rotation
from the pose output by Micro-GPS, we again mark the result as a failure. 
Finally, given a sequence of consecutive poses that should be temporally
coherent, we detect whether the poses of any frames differ significantly
from their neighbors.

The performance of our system, implementing the pipeline described in
Section~\ref{sec:online_localization}, is shown in
Table~\ref{tab:microgps}.  
The second column shows the elapsed time between capture of database and
test sequence, which demonstrates that our system is robust against changes
in the scene. The two columns at right show performance with 8-dimensional and
16-dimensional descriptors, respectively.  In the following, unless
otherwise specified, we use 16-dimensional descriptors due to their good
performance across all textures.

The correct pose is recovered
over 90\% of the time for most datasets (with independent per-frame
localization and no use of temporal coherence), with the exception of the
wood floor.  This is because relatively few SIFT features are available in
this dataset. (We have recently demonstrated that a highly repeatable feature
detector can be learned in a texture-specific manner~\cite{zhangdetector}.
Unfortunately, such a pipeline is currently not efficient enough for a mobile
platform.)

\subsection{Impact of Design Decisions}
\label{sec:impact}

\begin{figure}
  \centering
  \begin{tabular}{@{\hspace{0mm}}c@{\hspace{2mm}}c@{\hspace{0mm}}}
  \includegraphics[width=0.48\linewidth]{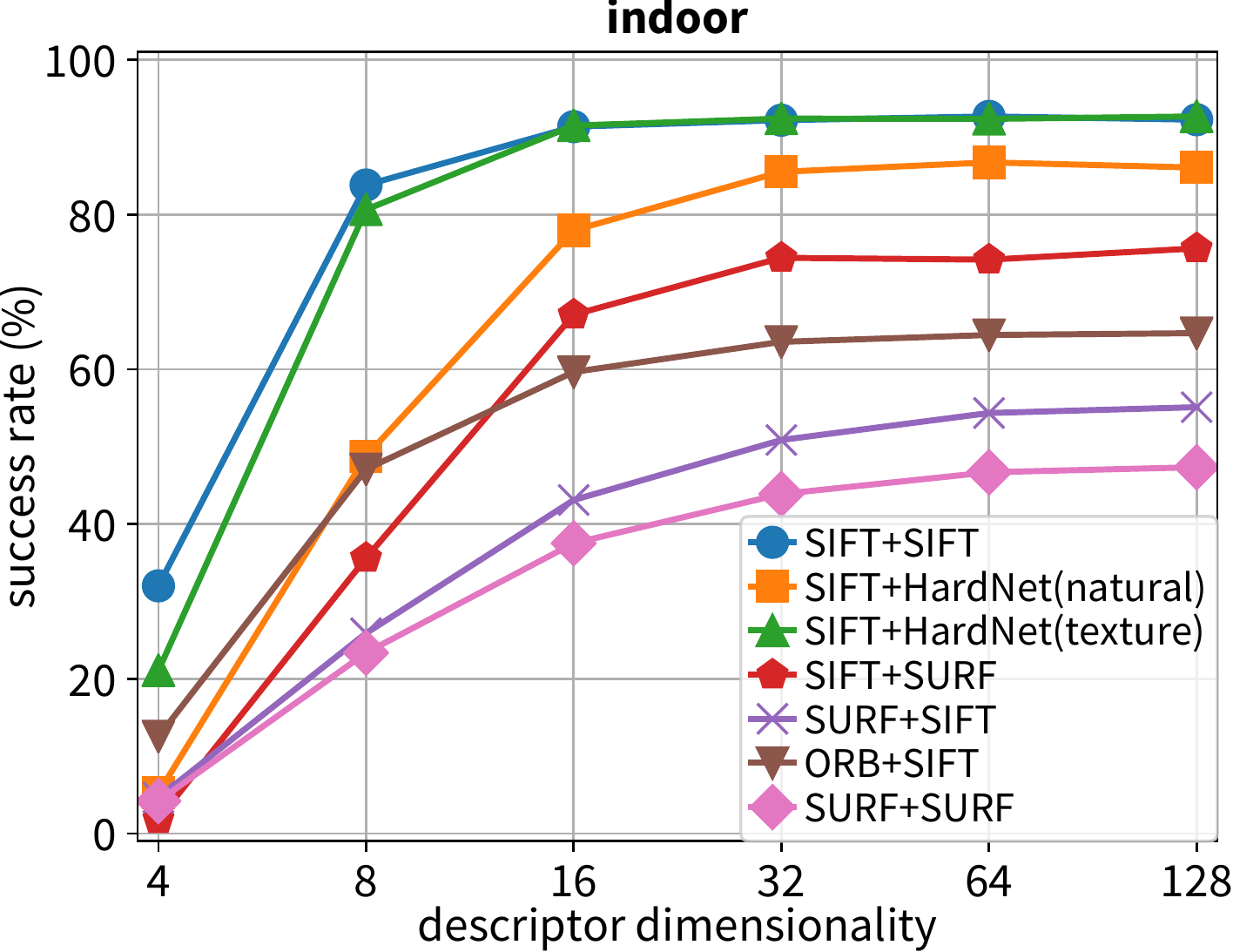} &
  \includegraphics[width=0.48\linewidth]{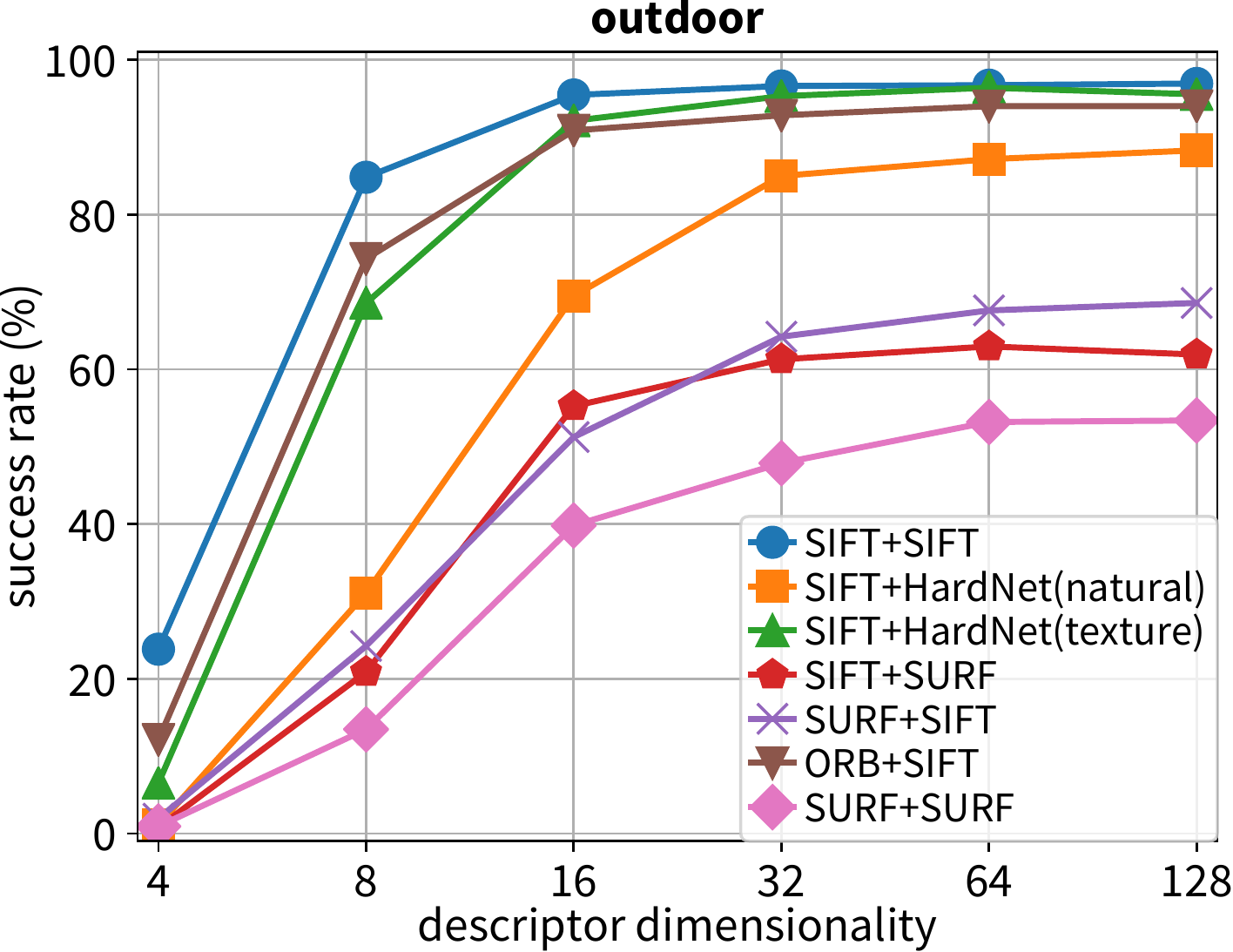}
  \end{tabular}
  \caption{
    The average performance of different \emph{detector} +
    \emph{descriptor} combinations on both indoor and outdoor datasets.  The
    horizontal axis indicates the dimensionality of descriptors after PCA.
  }
  \label{fig:compare_descriptors}
\end{figure}

{\noindent \bf Selection of Feature:} 
We first evaluate the impact on
accuracy of using different combinations of feature detector and
descriptor. While SIFT~\cite{lowe2004distinctive} has been popular since
its introduction more than a decade ago, more recent alternatives such as
SURF~\cite{bay2006surf}, ORB~\cite{rublee2011orb} have been shown to achieve
similar performance at lower computational cost. 
Even more recently, convolutional neural networks have been used in learned
descriptors~\cite{tian2017l2, mishchuk2017working, he2018local,
zhang2017learning, luo2018geodesc, keller2018learning} to achieve much
better matching performance than SIFT. In all of these cases, however, the
performance has been optimized for \emph{natural}, rather than
texture-like, images. To evaluate the effectiveness of a learned descriptor
on texture images, we select the recent well-performing
HardNet~\cite{mishchuk2017working} as our backbone network and learn a
texture descriptor (HardNet-texture) using patches cropped from our
dataset. During training, we also perform non-uniform intensity scaling to
account for possible changes in exposure. 
Figure~\ref{fig:compare_descriptors} compares the accuracy of different
combinations of feature detector and descriptor.  The SIFT \emph{detector}
outperforms both SURF and ORB, while both SIFT and HardNet-texture perform
better than alternative \emph{descriptors}.  Because the SIFT descriptor
can withstand more aggressive dimension reduction, it is the best choice
for our current deployment.  However, we observe that HardNet-texture shows
significant improvement compared to the original HardNet optimized for natural
images~\cite{winder2007learning}.  
This suggests that domain-specific training may hold the promise
for future improvements in the quality of learned descriptors.

\begin{table}
  \centering
  \caption{For each type of texture, we evaluate the success rate (in percentage) by
  running our system with PCA bases computed from each texture and the
  union of all textures. Best numbers are bolded.}
  \label{tab:pca_basis_cross}
  \setlength\tabcolsep{2pt}
  \def\colw{0.33in}
  \begin{tabularx}{\hsize}{
    >{\arraybackslash}p{\colw}
    >{\centering\arraybackslash}p{\colw}
    >{\centering\arraybackslash}p{\colw}
    >{\centering\arraybackslash}p{\colw}
    >{\centering\arraybackslash}p{\colw}
    >{\centering\arraybackslash}p{\colw}
    >{\centering\arraybackslash}p{\colw}
    >{\centering\arraybackslash}p{\colw}
    >{\centering\arraybackslash}p{\colw}}
    \toprule
    & \multicolumn{8}{c}{\scriptsize\bfseries Basis} \\
          {\scriptsize\bfseries Texture}
          & asphalt 
          & carpet
          & coarse
          & concrete
          & granite
          & tiles
          & wood
          & union \\
    \midrule
    asphalt&     95.24&      95.39&      95.24&      94.32&      95.08&  \bf 95.70&      94.32&      94.47\\ 
    carpet& \bf 100&      99.92& \bf 100& \bf 100& \bf 100&      99.92& \bf 100& \bf 100\\ 
    coarse&      99.09&  \bf 99.35&      99.09&  \bf 99.35&      99.09&  \bf 99.35&      98.83&      98.83\\ 
  concrete&      91.84&  \bf 93.73&      92.72&      93.35&      93.22&      93.22&      91.72&      92.85\\ 
   granite&  \bf 94.43&      93.85&      94.08&      93.62&  \bf 94.43&      94.08&      93.16&      93.97\\ 
     tiles&      98.27&  \bf 98.40&      97.90&      98.27&      98.09&  \bf 98.40&      97.59&      97.78\\ 
      wood&      76.85&  \bf 78.78&      78.14&      77.81&      77.81&  \bf 78.78&      77.49&      77.49\\ 
    \bottomrule
  \end{tabularx}
\end{table}

{\noindent \bf Choice of PCA Basis:}
We next investigate whether the PCA basis used for dimensionality reduction
should be specific to each dataset, or whether a ``universal'' basis
computed from the union of different textures achieves similar
performance.  Table \ref{tab:pca_basis_cross} shows localization performance
for each combination of texture and PCA basis, including a basis computed
from the union of all datasets. 
The difference caused by switching PCA basis is negligible, and we conclude
that there is no drawback of using a single PCA basis computed from all of the
datasets.

\subsection{Downward- vs. Outward-Facing Cameras}

Many existing systems focus on localization in natural images captured by
an outward-facing camera.  To compare our system with this approach to
localization, we add an outward-facing camera (identical to the one used
for Micro-GPS) to our setup
and we trigger both cameras simultaneously. We use the
VisualSFM structure-from-motion system~\cite{wu2011visualsfm,
wu2011multicore} to recover the 3D trajectory of the outward-facing camera.
In order to compare the resulting trajectory to ours, we project it
onto 2D (which we do by estimating the plane that the trajectory lies in
using least squares), and recover the relative global scale, rotation,
and translation (which we do by minimizing least-squares distance between
points taken at the same timestamps).

\begin{figure}
  \def\imw{0.52\hsize}
  \vspace{-5pt}
  \hbox to \hsize{\hskip-1pt
    \includegraphics[width=\imw]{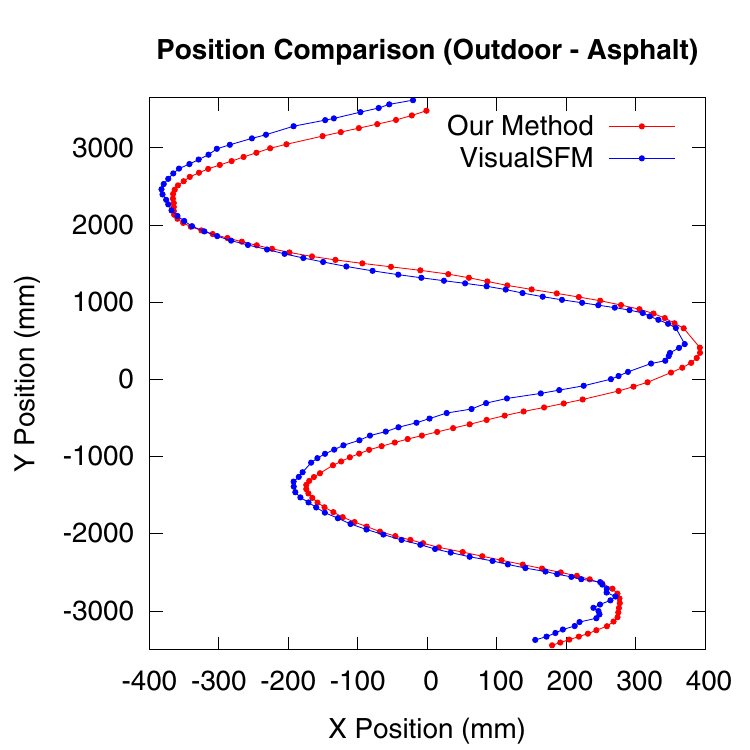}\hss
    \includegraphics[width=\imw]{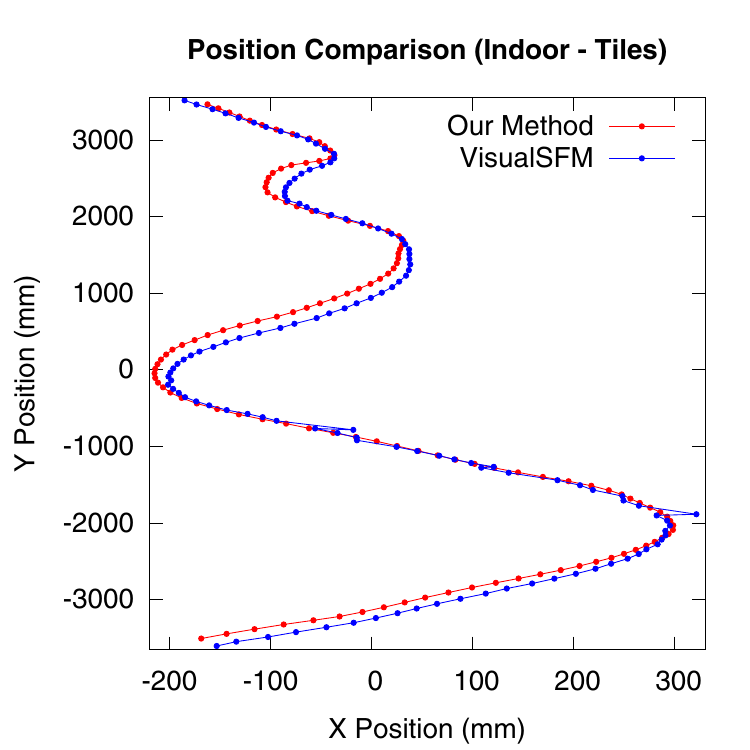}\hskip-6.5pt}
  \caption{
    Comparison of camera trajectories obtained using our system with
    downward-facing cameras (red lines) and a state-of-the-art 
    structure-from-motion system using outward-pointing cameras (blue lines). 
    Left: trajectories on the outdoor \emph{asphalt} dataset.
    The distance between the trajectories is 98.8~mm on average
    (maximum 211.5~mm) while the mean angle between camera poses is
    0.5~degrees (maximum 1.3~degrees).
    Right: trajectories on the indoor \emph{tiles} dataset.
    The mean distance is 62.9~mm (maximum 197.7~mm) and 
    the mean anglular difference is 2.2~degrees (maximum 2.5~degrees).
  }
  \label{fig:micro_gps_vs_visualsfm}
\end{figure}

Figure~\ref{fig:micro_gps_vs_visualsfm} compares both trajectories for two
different environments, outdoors (with asphalt ground texture) and indoors
(with the ``tiles'' texture).
We observe that the trajectory of VisualSfM (in blue) is much noisier 
than that of Micro-GPS, with discrepancies of many centimeters.  In contrast,
the difference in estimated orientations is small (usually below 1 degree),
suggesting that both methods were able to recover orientation successfully.

One factor that critically affects the performance of both systems is the
number of SIFT features that can be detected.  Our downward-facing camera
detects an average of 1319 features per frame in the (outdoor) asphalt
sequence and 2114 features in the (indoor) tile sequence, while the
outward-facing camera detects only 637 and 256 features in the same
settings.  More detected features typically leads to more matched features,
and hence greater localization accuracy.
Nevertheless, outward-facing cameras are commonly used for tracking and, at
the same speed of motion, are less susceptible to motion blur than
downward-facing cameras.  We conclude that our system can be used in
conjunction with a tracking system based on an outward-facing camera, with
comparable additional hardware and software costs. 

\subsection{Robustness}
\label{sec:robustness}


In the following section, we investigate the real-world applicability 
of the proposed system by stress-testing.

\begin{figure}
  \centering
  \includegraphics[width=\hsize]{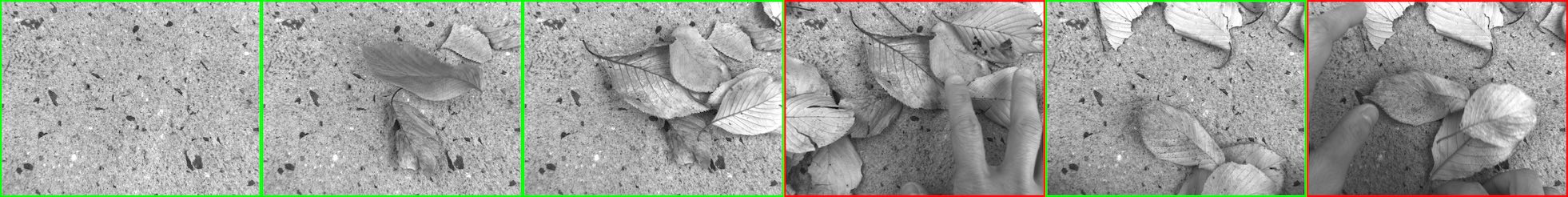}
  \includegraphics[width=\hsize]{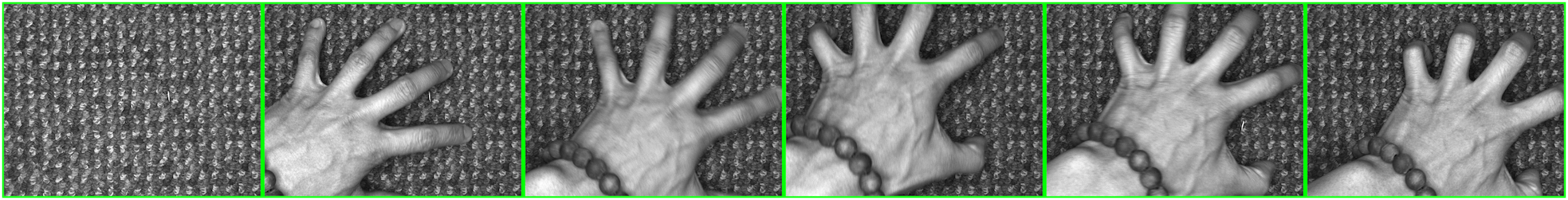}
  \caption{Introducing occlusion and perturbation. 
  First row: introducing occlusion by adding leaves.
  Second row: introducing occlusion and perturbation by scratching.
  The green bounding box represents success while red represents failure.}
  \label{fig:occlusion_test}
\vspace{0.1in}
  \centering
  \includegraphics[width=\hsize]{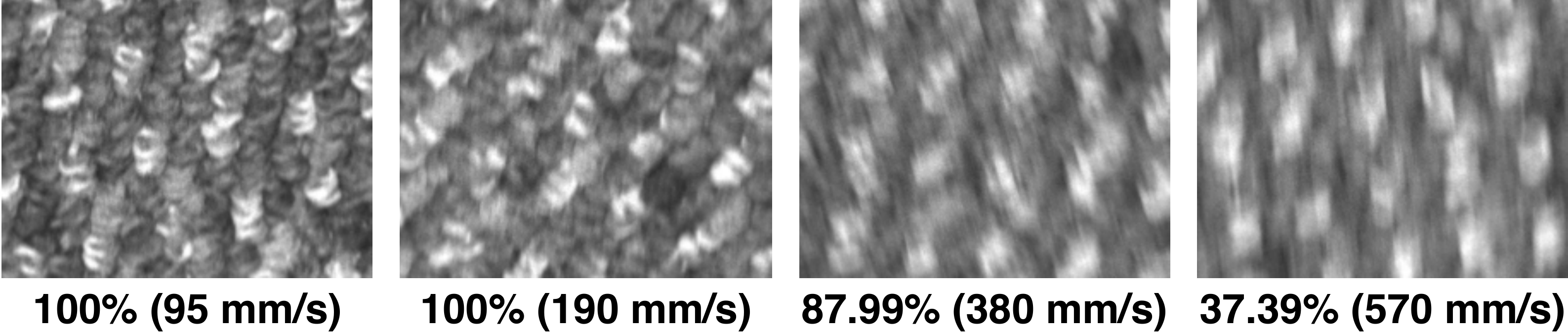}
  \caption{Examples of motion-blurred images captured on the carpet.
  The actual speed and success rate correspond are shown below.}
  \label{fig:vary_speed_success_rate}  
\end{figure}

{\noindent \bf Occlusion and Perturbation:}
\label{sec:occlusion_perturbation}
Two particular ways in which ground texture can change over time are
occlusion (e.g., dirt, leaves, etc.) and perturbation of soft materials
(e.g., walking on carpet).  
Figure \ref{fig:occlusion_test}, top, shows frames from a sequence in which 
more and more leaves are piled on a patch of concrete.  Frames outlined in
green represent success, while frames outlined in red represent failure
of localization.  Note that our voting procedure is robust against a
substantial amount of occlusion.
At bottom, we show frames from a sequence in which we scratch the carpet
by hand.  All frames in this sequence resulted in successful localization.

{\noindent \bf Motion Blur:}
Motion blur can easily happen when there is vibration or the camera is
moving too fast.  This perturbs both the feature detection and the computed
feature descriptors, making localization less accurate.  To evaluate how
motion blur can affect the performance of our system, we use a robot which
can run at a roughly-constant speed and evaluate performance by varying the
speed.
Unlike the previous experiments, we deliberately lower the shutter speed
to introduce motion blur.  
As shown in Figure~\ref{fig:vary_speed_success_rate}, our system can still
achieve reasonable success rate under moderate motion blur 
even with only 16-dimensional descriptors.
\begin{figure}
  \centering
  \includegraphics[width=0.98\linewidth]{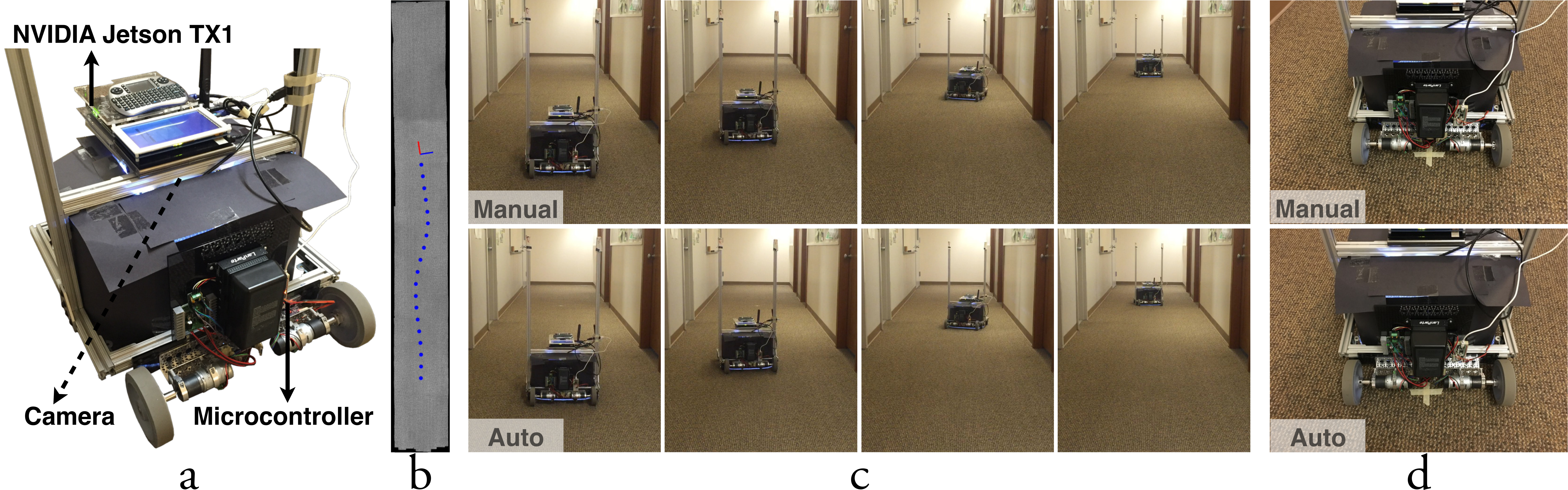}
  \caption{A demonstration of path following.
  (a) Micro-GPS is implemented as a component of a mobile robot.
  (b) We generate a path by manually driving the robot.
  (c) We then use Micro-GPS to repeat the path.  Screen-shots captured under
  manual and automatic driving modes are highly consistent.
  (d) The robot reaches the same ending position with high accuracy.
  }
  \label{fig:robot_path_following}
\end{figure}

\section{Application: Automatic Path Following}
\label{sec:follow}

Our system provides a simple, inexpensive solution to achieve fine absolute
positioning, and mobile robots having such a requirement represent an ideal
application.  
To demonstrate the practicality of this approach, we build a robot that is
able to follow a designed path exactly without any initialization of the
position.  Our robot (shown in Figure~\ref{fig:robot_path_following}a) has a
differential drive composed of two 24~V DC geared motors with encoders for
closed-loop control of the motors. 
Using the encoder readings, we implemented dead-reckoning odometry on board
to track the position of the robot at reasonable
accuracy within a short distance. The drift in odometry is corrected using
Micro-GPS running on the on-board NVIDIA Jetson TX1 computer at $\sim$4fps.


To test the repeatability of navigation using this strategy, we first
manually drive the robot along a particular path (Figure~\ref{fig:robot_path_following}b), and mark its final location
on the ground using a piece of tape.  
The robot then goes back to its starting position and re-plays the same
path, fully automatically.
The sequences of
the manual driving and automatic re-play are shown in the accompanying
video; screen-shots from the video are compared in Figure~\ref{fig:robot_path_following}c.  
As shown in Figure~\ref{fig:robot_path_following}d, the robot ends up in 
almost exactly the same position after automatic path following as it did 
after the manual driving.

\section{Discussion}
\label{sec:discussion}
To accommodate larger working areas, we would need to increase the volume of
the database, which could degenerate the robustness of our system due to
noisier feature matching. Also, performing matching within a large database
could raise the issue of efficiency. However, our system could work together
with existing systems that provide coarse localization (e.g.,
GPS) to narrow down the working area. 
While we believe that our system is applicable to a wide range of floor and
ground materials, localization is currently only possible in the presence
of unique and stable textures.  
When a reliable ground surface is absent, leveraging subsurface
features, such as what has been demonstrated in
LGPR~\cite{cornick2016localizing}, could be a feasible solution.

\def\bibfont{\footnotesize}
\bibliographystyle{IEEEtranN}
\bibliography{root} 

\end{document}